\documentclass[5p, times]{elsarticle}

\usepackage{lineno,hyperref}

\usepackage{xcolor}
\usepackage{amsmath,amssymb}
\usepackage{graphicx}
\usepackage{subcaption}
\usepackage{multicol}
\usepackage{tabularx}
\usepackage{float, caption, booktabs, cellspace}
\usepackage{soul}
\usepackage{ulem}

\modulolinenumbers[5]

\journal{Journal of \LaTeX\ Templates}

\begin{document}

\begin{frontmatter}

\title{An Efficient Framework for Visible-Infrared Cross Modality Person Re-Identification}


\author[mymainaddress]{Emrah~Basaran\corref{mycorrespondingauthor}}
\cortext[mycorrespondingauthor]{Corresponding author}
\ead{basaranemrah@itu.edu.tr}
\author[mysecondaryaddress]{Muhittin~G\"{o}kmen}
\author[mymainaddress]{Mustafa E.~Kamasak}

\address[mymainaddress]{Istanbul Technical University, Computer Engineering Department, Maslak, 34467, Istanbul, Turkey}
\address[mysecondaryaddress]{MEF University, Computer Engineering Department, Maslak, 34396, Istanbul, Turkey}




\begin{abstract}

Visible-infrared cross{\color{black}-}modality person re-identification (VI-ReId) is an {\color{black}essential} task for video surveillance in poorly illuminated or dark environments. 
Despite many recent studies on person re-identification in {\color{black}the} visible domain (ReId), there are few studies dealing {\color{black}specifically} with VI-ReId. 
Besides challenges that are common for both ReId and VI-ReId such as pose/illumination variations, background clutter and occlusion, VI-ReId has additional challenges as color information is not available in infrared images.
As a result, the performance of VI-ReId systems is typically lower than {\color{black}that of} ReId systems.
In this work, we propose a {\color{black}four}-stream framework to improve VI-ReId performance. We train a separate deep convolutional neural network in each stream using different representations of input images.
We expect that different and complementary features can be learned from each stream. In our framework, grayscale and infrared input images are used to train the ResNet in the first stream. In the second stream, RGB and  {\color{black}three}-channel infrared images (created by repeating {\color{black}the} infrared channel) are used. In the remaining two streams, we use local pattern maps as input images. These maps are generated utilizing local Zernike moments transformation. Local pattern maps are obtained from grayscale and infrared images in the  {\color{black}third} stream and from RGB and {\color{black}three}-channel infrared
images in the last stream. We improve the performance of the proposed framework by employing a re-ranking algorithm for post{\color{black}-}processing.
Our results indicate that the proposed framework outperforms current state-of-the-art
{\color{black}with a large margin by improving Rank-1/mAP by $29.79\%/30.91\%$ on SYSU-MM01 dataset, and by $9.73\%/16.36\%$ on RegDB dataset.}


\end{abstract}


\end{frontmatter}


\section{Introduction}\label{sec:introduction}

Person re-identification (ReId) can be defined as the retrieval of the images of a person from a gallery set{\color{black}, where the gallery and query sets consist} of the images captured by different cameras with different field-of-views. It is an {\color{black}essential} problem for various real-world scenarios, especially for security. Therefore, this problem has recently attracted {\color{black}considerable} attention of researchers working in {\color{black}the fields} of computer vision and machine learning. 

The illumination conditions, visible body parts, occlusion{\color{black},} and background complexity can vary extremely  {\color{black}in} images captured by different cameras. Due to the dynamic nature of the scene, these conditions can also change  {\color{black}in} images recorded by the same camera. These challenges and typically low{\color{black}-}resolution images make ReId a challenging computer vision problem. 
In the literature {\color{black}on this subject}, many different methods have been proposed in the last decade  {\color{black}that} address ReId from  {\color{black}various} aspects. 
The vast majority of the methods have been developed considering only the images captured by visible cameras. However, in dark or poorly illuminated environments, visible cameras cannot capture features that  {\color{black}can} distinguish people.

Most of the surveillance cameras used at night or in the dark usually operate in infrared mode in order to cope with poor illumination. Therefore, matching the person images captured by visible and infrared cameras is an important issue for video surveillance or miscellaneous applications. This issue is studied in literature as visible-infrared cross-modality person re-identification (VI-ReId) \cite{wu2017rgb,ye2018visible,ye2018hierarchical,wanglearning,kang2019person}. VI-ReId is the problem of retrieving the images of a person from a gallery set consisting of RGB (or infrared) images{\color{black},}
given an infrared (or RGB) query image.
{\color{black}For ReId,} one of the most important cues {\color{black}in} person images is obtained from the color. Therefore, the lack of color information in infrared images 
makes VI-ReId a very challenging problem.

In this paper, we show that ResNet \cite{he2016deep} architectures trained with the use of RGB and infrared images together can outperform the current state-of-the-art. These architectures are widely used for image classification and other computer vision problems. They can learn the common feature representations for RGB and infrared images of the same individual as well as the {\color{black}distinctive} properties between the individuals better than the existing methods proposed for VI-ReId. In this study, we introduce a {\color{black}four}-stream framework built with ResNet architectures. There is no weight sharing between the ResNets in the framework, and in order to obtain different and complementary features as much as possible, we train each of them using a different representation of input images. In the first stream, RGB images converted to grayscale and infrared images are used together. In this way, features are extracted using only the shape and pattern information. In the second stream, to take advantage of the color information, RGB images and the {\color{black}three}-channel infrared images (obtained by repeating the infrared image) are used. As mentioned above, one of the most important cues for ReId is obtained from the color information. However, due to the lack of color in infrared images, local shape and pattern information is critical for VI-ReId. Therefore, in addition to the features obtained from the raw (RGB, grayscale and infrared) images in the first two streams of the framework, in the last two streams, features are {\color{black}derived} from the local pattern maps. 
In the {\color{black}third} stream, we first apply the local Zernike moments (LZM) transformation \cite{Sariyanidi2012} on grayscale and infrared images to expose the local patterns in the images. With the LZM transformation, different number{\color{black}s} of local pattern maps are generated by computing the Zernike moments around each pixel. We train the ResNet in the {\color{black}third} stream by using the LZM pattern maps of grayscale and infrared images. In the last stream of the framework, we generate the pattern maps by exposing the local patterns separately in R, G{\color{black},} and B channels. 


The contributions of this paper are summarized as follows:
\begin{itemize}
\item We propose a novel framework {\color{black}that} consists of {\color{black}four} streams. In order to obtain complementary features from each stream, we train a ResNet architecture in each stream by using a different representation of input images.

\item This {\color{black}work} is the first study employing the LZM transformation for ReId and training a deep convolutional neural network with the LZM pattern maps. 

\item Our framework\footnote{The project webpage: https://github.com/emrahbasaran/cross-re-id} {\color{black} outperforms current state-of-the-art with a large margin by improving Rank-1/mAP by $29.79\%/$ $30.91\%$ on SYSU-MM01 dataset, and by $9.73\%/16.36\%$ on RegDB dataset. }
\end{itemize}


\section{Related Work}
\label{sec:related_work}

In recent years, deep convolutional neural networks (DCNN) have led to {\color{black}significant} progress in many different computer vision areas. Person re-identification (ReId) is one of the challenging computer vision problems. {\color{black}In} order to cope with {\color{black}issues} such as illumination conditions, variations in pose, closure{\color{black},} and background clutter, researchers have developed many DCNN frameworks tackling ReId in different ways \cite{mogelmose2013tri,chen2017person,sarfraz2017pose,kalayeh2018human,deng2018image}.

In {\color{black}the} majority of the studies on ReId, only the images in {\color{black}the} visible spectrum (RGB-based) are taken into account when addressing the challenges, and the developed methods are intended for RGB-based images only. There are only a few studies for visible-infrared or visible-thermal
cross{\color{black}-}modality person re-identification. However, they are important issues for video surveillance or miscellaneous applications that have to be performed in poorly illuminated or dark environments. 
In \cite{wu2017rgb}, the authors have analyzed different network structures {\color{black}for VI-ReId}, including one-stream and two-stream architectures and asymmetric full{\color{black}y}-connected layer. {\color{black}They have} found the performance of the one-stream architecture to be better. 
In addition, they have proposed deep zero padding to contribute to the performance of the one-stream network. Since the authors train the architectures using grayscale input images, they do not take advantage of the color of RGB images. 
Kang \textit{et al.} \cite{kang2019person} propose a one-stream model for cross-modality re-identification. They create single input images by placing visible and infrared images in different channels or by concatenating them. Using these input images consisting of positive and negative pairs, they train a DCNN employing {\color{black}two}-class classification loss. In the study, some pre-processing methods are also analyzed.
Ye \textit{et al.} \cite{ye2018visible} use a {\color{black}two}-stream DCNN to obtain modality-specific features from the images. One stream of the network is fed with RGB while the other is fed with infrared (or thermal) images, and on top of the streams, there is a shared fully connected layer in order to learn a common embedding space. The authors train the  {\color{black}two}-stream model using multi-class classification loss along with a ranking loss proposed by them. A very similar  {\color{black}two}-stream model is employed in \cite{ye2018hierarchical}, but this model is trained utilizing contrastive loss instead of ranking loss used in \cite{ye2018visible}. To improve {\color{black}the} discrimination ability of the features extracted by {\color{black}the}  {\color{black}two}-stream model, they propose a hierarchical cross-modality metric learning. In this method, modality-specific and modality-shared metrics are jointly learned in order to minimize cross{\color{black}-}modality inconsistency. 
In \cite{dai2018cross}, a generative adversarial network (GAN) is proposed for visible-infrared re-identification. The authors use a DCNN, which is trained employing both multi-class classification and triplet losses, as {\color{black}a} generator to extract features from RGB and infrared images. The discriminator is constructed as a modality classifier to distinguish between RGB and infrared representations. Another work utilizing a variant of GAN has introduced by Kniaz \textit{et al.} \cite{kniaz2018thermalgan} for visible-thermal re-identification. In their method, a set of synthetic thermal images are generated for an RGB probe image by employing a GAN framework. Then, the similarities between the synthetic probe and the thermal gallery images are calculated. Wang \textit{et al.} \cite{wanglearning} propose a network dealing {\color{black}with} the modality discrepancy and appearance discrepancy separately. They first create a unified multi-spectral representation by projecting the visible and infrared images to a unified space. Then, they train a DCNN using the multi-spectral representations by employing triplet and multi-class classification losses. There are also some other works \cite{mogelmose2013tri, nguyen2017person} trying to improve RGB-based ReId by fusing the information from different modalities.

In the literature, most of the existing studies on cross-modality matching or retrieval have been performed for heterogeneous face recognition (HFR) and text-to-image (or image-to-text) matching in the past decade. Studies on HFR, which are more relevant to VI-ReId, can be reviewed in three perspectives \cite{he2019cross}: image synthesis, latent subspace, and domain-invariant features. In the first group, before the recognition, face images are transformed  {\color{black}into} the same domain \cite{song2018adversarial, lezama2017not, huang2017beyond}. The approach followed by the studies in the second group is the projection of the data of different domains into a common latent space \cite{lei2012coupled, kan2016multi, jin2017multi}. In the recent works exploring domain-invariant feature representations, deep learning based methods are proposed. In \cite{saxena2016heterogeneous}, the features are obtained from a DCNN pre-trained on visible spectrum images, and metric learning methods are used to overcome inconsistencies between the modalities. He \textit{et al.} \cite{he2018wasserstein} build a model in which the low-level layers are shared{\color{black},} and the high-level layers are divided into infrared, visible{\color{black},} and shared infrared-visible branches. Peng \textit{et al.} \cite{peng2019dlface} propose to generate features from facial patches and utilize a novel cross-modality enumeration loss while training the network. In \cite{deng2019mutual}, Mutual Component Analysis (MCA) \cite{li2016mutual} is integrated as a fully-connected layer into DCNN, and an MCA loss is proposed.

In this study, we propose a framework that utilizes the pattern maps generated by local Zernike moments (LZM) transformation \cite{Sariyanidi2012}. {\color{black}As robust and holistic image descriptors,} Zernike moments (ZM) \cite{teague1980image} are commonly used in many different computer vision problems, i.e., character \cite{Kan2002}, fingerprint \cite{zhai2010application} and iris \cite{tan2014accurate} recognition, and object detection \cite{Bera2019}. LZM transformation is proposed by Sariyanidi \textit{et al.} \cite{Sariyanidi2012} to utilize ZM at {\color{black}the} local scale for shape/texture analysis. In this transformation, images are encoded by calculating the ZM around each pixel. Thus, the local patterns in the images are exposed{\color{black},} and a rich representation is obtained. In literature, LZM transformation is used in various studies such as face recognition \cite{Sariyanidi2012,alasag2014face, basaran2018efficient, basaran2014efficient}, facial expression \cite{fan2017dynamic, gaziouglu2017facial} and facial affect recognition \cite{sariyanidi2013local}, traffic sign classification \cite{bacsaran2013traffic}, loop closure detection \cite{sariyanidi2013loop, erhan2016patterns}, {\color{black}and} interest point detector \cite{ozbulak2015rotation}.
In these works, the features are obtained directly from the LZM pattern maps. 
Unlike them, in this study, we use the LZM pattern maps as input images to train DCNNs.
This is the first study training DCNNs with the LZM pattern maps. {\color{black}However, there have been} some attempts employing ZM in convolutional networks. Mahesh \textit{et al.} \cite{mahesh2017invariant} initialize the trainable kernel coefficients by utilizing ZM with different moment orders. In \cite{yoon2016semi} and \cite{wu2017momentsnet}, learning-free architectures are built constructing the convolutional layers by using ZM. Sun \textit{et al.} \cite{sun2018zernet} introduce a novel concept of Zernike convolution to extend convolutional neural networks for 2D-manifold domains.


\section{Proposed Method}
\label{sec:proposed_method}

In this section, we begin by giving the details of ResNet \cite{he2016deep} architectures. Then, we describe the LZM transformation \cite{Sariyanidi2012} and introduce our {\color{black}four}-stream framework proposed for VI-ReId. Finally, we explain the ECN re-ranking algorithm \cite{sarfraz2017pose} used as post-processing.

\subsection{ResNet Architecture}
\label{sec:resnet}

We show the performance of our proposed framework by using {\color{black}three} different ResNet architectures which have different depths. These architectures have 50, 101{\color{black},} and 152 layers (we call these architectures as ResNet-50, ResNet-101{\color{black},} and ResNet-152, respectively, in the rest of the paper), and the{\color{black}ir} details are provided in Table \ref{table:resnet_layers}. 
\begin{table}
\scriptsize
\centering
\caption{ResNet architectures used in this study. The building blocks are given in brackets. {\color{black}Each} row in the brackets indicates the kernel sizes and the number of output channels of convolution layers used in the building blocks.}
\begin{tabular}{ccc}
\hline
\multicolumn{1}{c}{50-layer}   & \multicolumn{1}{c}{101-layer}  & 152-layer  \\ \hline
\multicolumn{3}{c}{7 x 7, 64, stride 2}                                        \\ \hline
\multicolumn{3}{c}{3 x 3 max pool, stride 2}                                   \\ \hline
\multicolumn{1}{c}{ 
                                    \Bigg[
                                      \begin{tabular}{ccc}
                                      $1\times1,64$ \\
                                      $3\times3,64$ \\
                                      $1\times1,256$ 
                                      \end{tabular}
                                    \Bigg]
                                               x 3} &
                \multicolumn{1}{c}{\Bigg[
                                      \begin{tabular}{ccc}
                                      $1\times1,64$ \\
                                      $3\times3,64$ \\
                                      $1\times1,256$ 
                                      \end{tabular}
                                    \Bigg] x 3} & 
                                    \Bigg[
                                      \begin{tabular}{ccc}
                                      $1\times1,64$ \\
                                      $3\times3,64$ \\
                                      $1\times1,256$ 
                                      \end{tabular}
                                    \Bigg] x 3 \\ \hline
\multicolumn{1}{c}{ 
                                    \Bigg[
                                      \begin{tabular}{ccc}
                                      $1\times1,128$ \\
                                      $3\times3,128$ \\
                                      $1\times1,512$ 
                                      \end{tabular}
                                    \Bigg]
                                               x 4} &
                \multicolumn{1}{c}{\Bigg[
                                      \begin{tabular}{ccc}
                                      $1\times1,128$ \\
                                      $3\times3,128$ \\
                                      $1\times1,512$ 
                                      \end{tabular}
                                    \Bigg] x 4} & 
                                    \Bigg[
                                      \begin{tabular}{ccc}
                                      $1\times1,128$ \\
                                      $3\times3,128$ \\
                                      $1\times1,512$ 
                                      \end{tabular}
                                    \Bigg] x 8 \\ \hline
\multicolumn{1}{c}{ 
                                    \Bigg[
                                      \begin{tabular}{ccc}
                                      $1\times1,256$ \\
                                      $3\times3,256$ \\
                                      $1\times1,1024$ 
                                      \end{tabular}
                                    \Bigg]
                                               x 6} &
                \multicolumn{1}{c}{\Bigg[
                                      \begin{tabular}{ccc}
                                      $1\times1,256$ \\
                                      $3\times3,256$ \\
                                      $1\times1,1024$ 
                                      \end{tabular}
                                    \Bigg] x 23} & 
                                    \Bigg[
                                      \begin{tabular}{ccc}
                                      $1\times1,256$ \\
                                      $3\times3,256$ \\
                                      $1\times1,1024$ 
                                      \end{tabular}
                                    \Bigg] x 36 \\ \hline
\multicolumn{1}{c}{ 
                                    \Bigg[
                                      \begin{tabular}{ccc}
                                      $1\times1,512$ \\
                                      $3\times3,512$ \\
                                      $1\times1,2048$ 
                                      \end{tabular}
                                    \Bigg]
                                               x 3} &
                \multicolumn{1}{c}{\Bigg[
                                      \begin{tabular}{ccc}
                                      $1\times1,512$ \\
                                      $3\times3,512$ \\
                                      $1\times1,2048$ 
                                      \end{tabular}
                                    \Bigg] x 3} & 
                                    \Bigg[
                                      \begin{tabular}{ccc}
                                      $1\times1,512$ \\
                                      $3\times3,512$ \\
                                      $1\times1,2048$ 
                                      \end{tabular}
                                    \Bigg] x 3 \\ \hline
\multicolumn{3}{c}{global average pool}   \\
\hline
\end{tabular}
\label{table:resnet_layers}
\end{table}
Each ResNet model shown in Table \ref{table:resnet_layers} starts with a $7\times7$ convolution layer followed by a $3\times3$ max pooling. The next layers of the models consist of {\color{black}a} different number of {\color{black}stacked} building blocks {\color{black}with} a global average pooling layer at {\color{black}their top}. In Table \ref{table:resnet_layers}, we show the building blocks in brackets where each row in the brackets indicates the kernel sizes and the number of output channels of convolution layers used in the building blocks. 

\subsection{LZM transformation}
\label{sec:zernike}

 Zernike moments (ZM) \cite{teague1980image} are commonly used to generate holistic image descriptors for various computer vision tasks \cite{Kan2002, zhai2010application, tan2014accurate}. Independent holistic characteristics of the images are exposed with the calculation of ZM of different orders \cite{chong2003comparative}. 
In order to reveal the local shape and texture information from the images by utilizing ZM, local Zernike moments (LZM) transformation is proposed in \cite{Sariyanidi2012}.

\begin{figure}
    \centering
    \includegraphics[width=8.8cm]{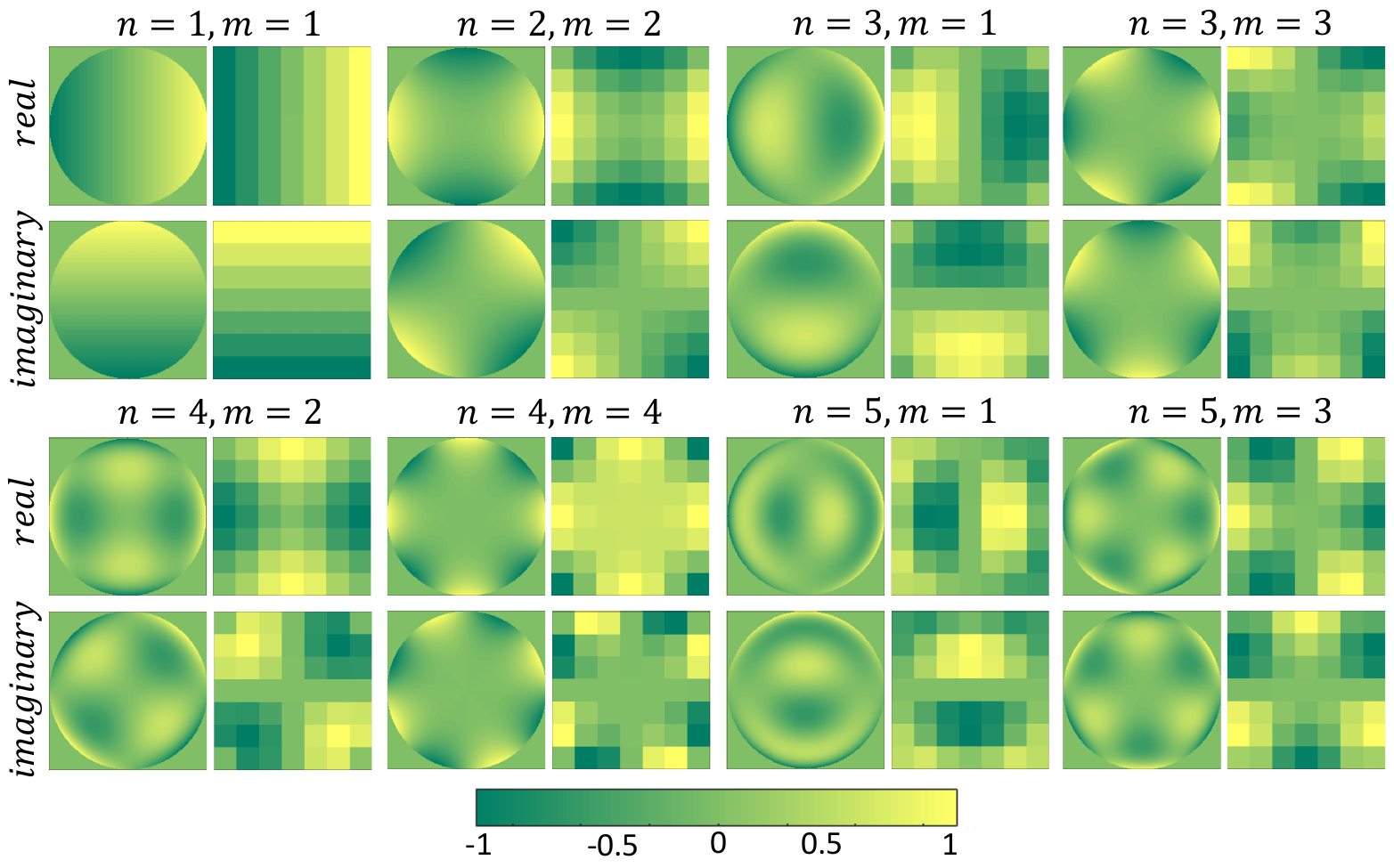}
    \caption{Zernike polynomials (odd columns){\color{black},} and the corresponding $7\times7$ LZM filters (even columns) generated using the values up to $n=5$, $m=3$.}
    \label{fig:filters}
\end{figure}

Zernike moments of an image are calculated using orthogonal Zernike polynomials \cite{teague1980image}. These polynomials are defined in polar coordinates within a unit circle and represented as follows:
\begin{equation}
\label{eqn:Vnm}
V_{nm}(\rho,\theta)=R_{nm}(\rho)e^{-\hat{j}m\theta},\quad\hat{j}=\sqrt{-1}
\end{equation}
where $R_{nm}(\rho)$ are real-valued radial polynomials, and $\rho$ and $\theta$ are the radial coordinates calculated as
\begin{equation}
\label{eqn:rho}
\rho_{ij}=\sqrt{x_i^2+y_j^2},
\end{equation}
\begin{equation}
\label{eqn:theta}
\theta_{ij}=tan^{-1}(y_j/x_i).
\end{equation}
Here, $x_i$ and $y_j$ are the pixel coordinates scaled to {\color{black}the} range of $[-1,1]$. $R_{nm}(\rho)$ polynomials are defined as
\begin{equation}
\label{eqn:Rnm}
R_{nm}(\rho)=\sum_{k=0}^{\frac{n-|m|}{2}}(-1)^k\frac{(n-k)!}{k!(\frac{n+|m|-2k}{2})!(\frac{(n-|m|-2k)}{2})!}\rho^{n-2k}.
\end{equation}
In (\ref{eqn:Vnm}) and (\ref{eqn:Rnm}), $n$ and $m$ are the moment order and the repetition, respectively. They take values such that $n-m$ is even, $0\leq n$ and $0\leq m\leq n$. 
Real and imaginary components of some $V_{nm}(\rho,\theta)$ Zernike polynomials are shown in Figure \ref{fig:filters}. 
Finally, ZM of an image $f(i,j)$ are calculated as:
\begin{equation}
\label{eqn:ZM}
	Z_{nm}= \frac{2(n+1)}{\pi (N-1)^2} \sum_{i=0}^{N-1}\sum_{j=0}^{N-1}V_{nm}(\rho_{ij},\theta_{ij})f(i,j),
\end{equation}

\begin{figure}
    \centering
    \includegraphics[width=8.6cm]{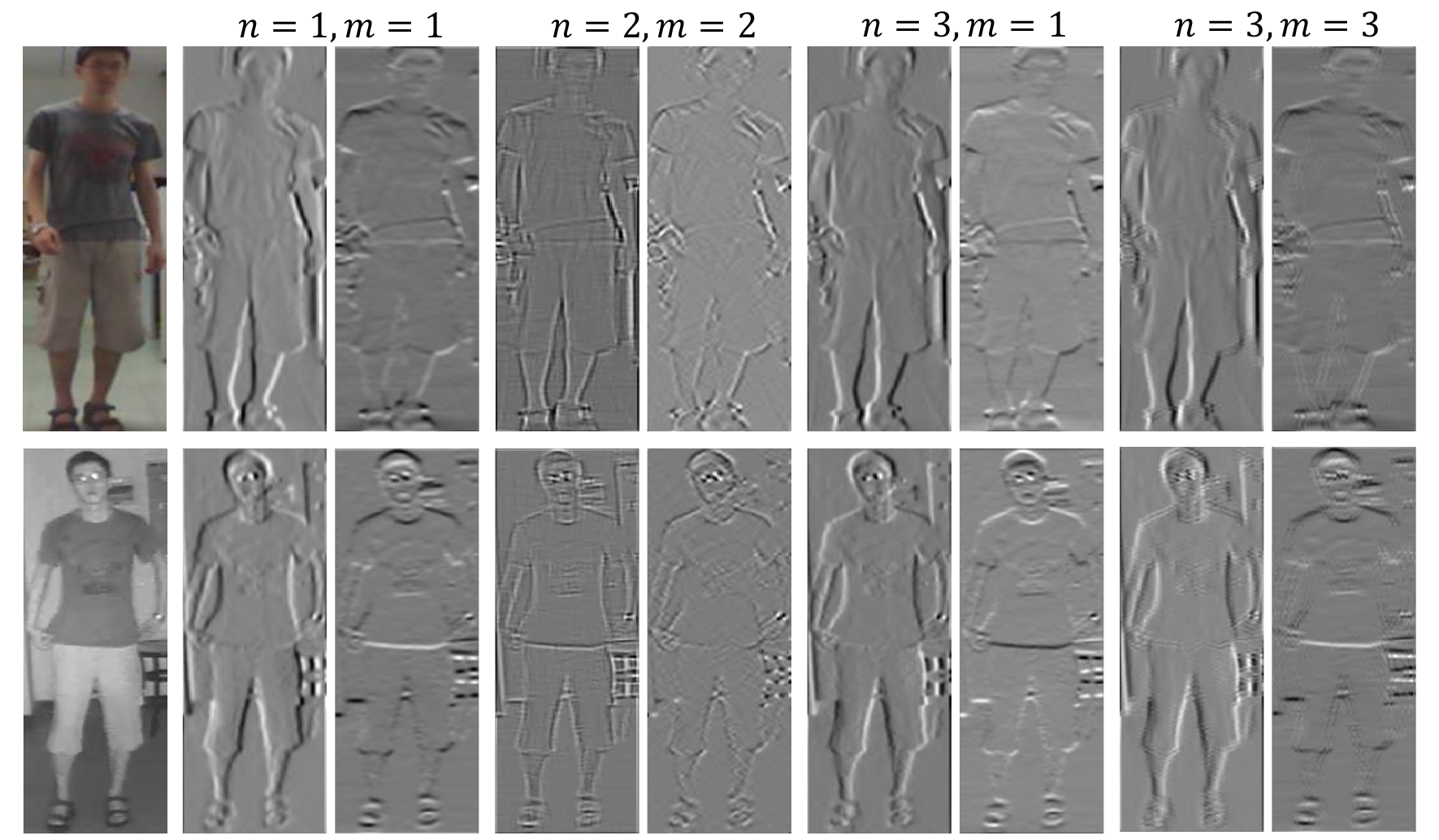}
    \caption{Pattern maps generated using the LZM filters given in Figure \ref{fig:filters} up to $n=3$. First column shows the RGB (first row) and infrared (second row) input images. Even and odd columns show the real and imaginary components, respectively. Pattern maps are normalized for better visualization.}
    \label{fig:lzm_outs}
\end{figure}

In the LZM transformation, images are encoded in $k\times k$ neighborhood of each pixel, as follows:
\begin{equation}
\label{eqn:lzm}
{Z}_{nm}^k(i,j)=\frac{2(n+1)}{\pi (k-1)^2}\sum_{p,q=-\frac{k-1}{2}}^{\frac{k-1}{2}}
{V}_{nm}^k(p,q)f(i-p,j-q)
\end{equation}
where ${V}_{nm}^k(p,q)$ represents the Zernike polynomials used as $k\times k$ filtering kernels and defined as
\begin{equation}
\label{eqn:Vk}
{V}_{nm}^k(i,j)=V_{nm}(\rho_{ij},\theta_{ij}).
\end{equation}
We give real and imaginary components of some ${V}_{nm}^k(i,j)$ filters in Figure \ref{fig:filters}. {\color{black}In} Figure \ref{fig:lzm_outs}, we show a few pattern maps generated using the LZM filters. The filters constructed using $m=0$ are not used in the LZM transformation since the imaginary components of Zernike polynomials are not generated. According to the moment order $n$, the number of complex filters is calculated using
\begin{equation}
\label{eqn:kn}
K(n)=\bigg\{\begin{matrix}
\frac{n(n+2)}{4} \text{\quad if $n$ is even},\\\frac{(n+1)^2}{4} \text{\quad if $n$ is odd.}
\end{matrix}
\end{equation}
Since we create a pattern map with each of the real and imaginary components of complex filters, the total number of local pattern maps becomes $2\times K(n)$.

\subsection{Person Re-Identification Framework}

\begin{figure*}[ht!]
    \centering
    \includegraphics[width=.9\textwidth]{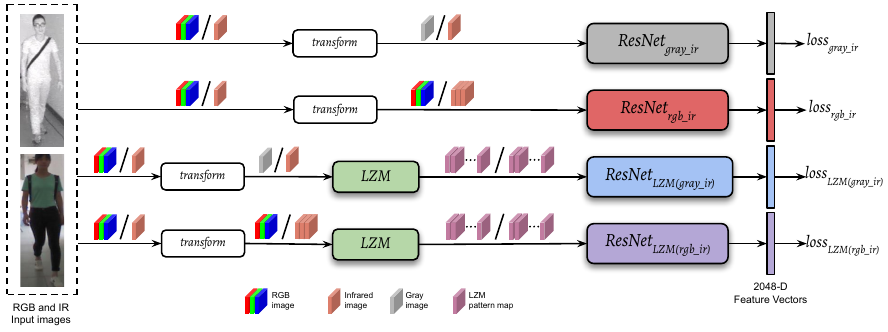}
    \caption{Our proposed VI-ReId framework. There is no weight sharing between ResNets in the framework, and each ResNet is trained with a different representation of input images. In the first stream, grayscale and infrared input images are used, while RGB and {\color{black}three}-channel infrared images (created by repeating infrared channel) are used in the second stream. In the other streams, LZM pattern maps are used as input images. These maps are obtained from grayscale and infrared images in {\color{black}the third} stream and separately from R, G{\color{black},} and B channels in {\color{black}the fourth} stream. 
    We employ multi-class classification with softmax cross-entropy loss for each stream separately during the training. In the evaluation phase, the final representations are constructed by concatenating the feature vectors obtained from each stream.
    Best viewed in color.}
    \label{fig:method}
\end{figure*}

The proposed framework for VI-ReId is shown in Figure \ref{fig:method}. This framework includes {\color{black}four} streams, and in each stream, a separate ResNet architecture is trained using both infrared and RGB images. Since each of {\color{black}the} ResNets accepts a different representation of input images, different and complementary features are obtained from each stream.

There is a large visual difference between infrared and RGB images since infrared images have a single channel with invisible light information. This visual difference is reduced by converting RGB images to grayscale in the upper stream of the proposed framework, and the training is performed using infrared and grayscale images. In this way, $ResNet_{gray\_ir}$ in the upper stream of the framework learns to extract robust features for the person images by utilizing only the shape and texture information.

For ReId, one of the most important cues for person images is obtained from the color. Therefore, in the second stream of the proposed framework, $ResNet_{rgb\_ir}$ is trained using RGB and {\color{black}three}-channel infrared (created by repeating the infrared channel) images. 
Unlike the one that is trained in the upper stream, the model trained in this stream uses color information of the RGB images and extracts discriminative features for the images in different modalities with visually large differences.

In VI-ReId, due to the lack of color in infrared images, the local shape and pattern information {\color{black}has} great importance while matching the infrared and RGB images. Therefore, in addition to the two streams mentioned above, there are two other streams in the proposed framework in order to obtain stronger features related to local shape and pattern information. In the third stream, first, the local patterns are exposed {\color{black}by} applying the LZM transformation on the grayscale and infrared images. Then, $ResNet_{LZM(gray\_ir)}$ is trained using the generated LZM pattern maps. As indicated in Section \ref{sec:zernike}, a number of complex pattern maps are obtained as a result of the LZM transformation. The input tensor $I_{LZM(gray\_ir)}$ for $ResNet_{LZM(gray\_ir)}$ is prepared by concatenating the real and imaginary components of the complex maps. $I_{LZM(gray\_ir)}$ is denoted as 
\begin{equation}
\label{eqn:lzm_gray}
I_{LZM(gray\_ir)}=T_{LZM}(I_{gray/ir})
\end{equation}
where $I_{gray/ir}$ represents the grayscale or infrared image and $T_{LZM}(I)$ is defined as
\begin{equation}
\label{eqn:lzm_tensor}
\begin{aligned}
T_{LZM}(I)=[re_{11}^k(I), im_{11}^k(I), re_{22}^k(I), im_{22}^k(I),\cdots, re_{nm}^k(I), im_{nm}^k(I)].
\end{aligned}
\end{equation}
In Eq. (\ref{eqn:lzm_tensor}), $re_{nm}^k(I)$ and $im_{nm}^k(I)$ are the real and imaginary components of  {\color{black}a} complex LZM pattern map{\color{black},} which is generated using the moment order $n$, the repetition $m${\color{black},} and the filter size $k$.
In the last stream of the proposed framework, LZM transformation is applied separately for the R, G and B channels, and the pattern maps obtained from these channels are concatenated to form the input tensor $I_{LZM(rgb\_ir)}$ for $ResNet_{LZM(rgb\_ir)}$, such as
\begin{equation}
\label{eqn:lzm_rgb}
\begin{aligned}
I_{LZM(rgb\_ir)}=[T_{LZM}(I_{rgb}^r), T_{LZM}(I_{rgb}^g), T_{LZM}(I_{rgb}^b)].
\end{aligned}
\end{equation}
Here, $I_{rgb}^r$, $I_{rgb}^g${\color{black},} and $I_{rgb}^b$ are the R, G{\color{black},} and B channels of the input image $I$. For infrared images, as mentioned earlier, {\color{black}the} infrared channel is repeated for R, G{\color{black},} and B, such as
\begin{equation}
\label{eqn:ir_rgb}
\begin{aligned}
I_{LZM(rgb\_ir)}=[T_{LZM}(I_{ir}), T_{LZM}(I_{ir}), T_{LZM}(I_{ir})].
\end{aligned}
\end{equation}

In order to train the ResNet architectures in the proposed framework, multi-class classification with softmax cross-entropy loss is employed. Thus, in addition to learning to extract common features for the images in different modalities, the ResNet models also learn to extract features that express differences between individuals. As shown in Figure \ref{fig:method}, the loss for each stream is calculated separately during the training. In the evaluation phase, the final representations for the person images are constructed by concatenating the feature vectors obtained from each stream and normalized using $\ell_{2}$-norm.

\subsection{Re-Ranking}

{\color{black}In recent studies \cite{sarfraz2017pose, ye2016person, zhong2017re}, it has been shown that re-ranking techniques have a significant contribution to the performance of person re-identification. Therefore, in this study, }  {\color{black}we} utilize Expanded Cross Neighborhood (ECN) re-ranking algorithm \cite{sarfraz2017pose} as a post-processing {\color{black}element} to further improve the performance. 
In this algorithm, the distance between a probe image $p$ and a gallery image $g_i$ from a gallery set $G$ with $B$ images $G=\{g_i|i=1,2,\cdots,B\}$ is defined as 
\begin{equation}
\label{eqn:ecn_dist}
\begin{aligned}
ECN(p,g_i)=\frac{1}{2M}\sum^M_{j=1}d(pN_j,g_i)+d(g_iN_j,p).
\end{aligned}
\end{equation}
Here, $pN_j$ and $g_iN_j$ are the $jth$ neighbors in the expanded neighbor sets $N(p,M)$ and $N(g_i,M)$ of the probe and $ith$ gallery images, respectively. $d(\cdot)$ represents the distance between the images, and $M$ is the total number of  {\color{black}neighbors} in a set. In order to construct the expanded neighborhood sets, first, an initial rank list $\mathcal{L}(p,G)=\{g^o_1,\cdots,g^o_B\}$ in increasing order is created for each image by calculating the pairwise  Euclidean distances between all images in the probe and gallery sets. Then, the expanded neighbor set $N(p, M)$ for a probe image $p$ is given as
\begin{equation}
\label{eqn:ecn_nset1}
\begin{aligned}
N(p,M)\leftarrow\{N(p,t),N(t,q)\}
\end{aligned}
\end{equation}
where the set $N(p,t)$ consists of the $t$ nearest neighbors of probe $p$, and the set $N(t,q)$ contains the $q$ nearest neighbors of each of the images in $N(p,t)$ such that:
\begin{equation}
\label{eqn:ecn_nset2}
\begin{aligned}
N(p,t)=\{g^o_i|i=1,2,\cdots,t\}\\
N(t,q)=\{N(g^o_i,q),\cdots,N(g^o_t,q)\}
\end{aligned}
\end{equation}
The expanded neighbor set $N(g_i,M)$ for gallery image $g_i$ is obtained in the same way.
In Equation \ref{eqn:ecn_dist}, as suggested by the authors of \cite{sarfraz2017pose}, we compute the distance between the pairs using a list comparison similarity measure proposed in \cite{jarvis1973clustering} and defined as
\begin{equation}
\label{eqn:ecn_list_dist}
\begin{aligned}
R(\mathcal{L}_i,\mathcal{L}_j)=\sum^B_{b=1}[K+1-pos_i(b)]_+\times[K+1-pos_j(b)]_+.
\end{aligned}
\end{equation}
In this equation, $[\cdot]_+=max(\cdot,0)$, $K$ is the number of nearest neighbors to be considered, and $pos_i(b)$ and $pos_j(b)$ show the position of image $b$ in the rank lists $\mathcal{L}_i$ and $\mathcal{L}_j$, respectively. The distance $d$ between the pairs is calculated as $d=1-R$ after scaling the range of the values of $R$ between $0$ and $1$. For the parameters $t$, $q${\color{black},} and $K$ of ECN re-ranking algorithm, we use the same setting given in \cite{sarfraz2017pose} such that $t=3$, $q=8${\color{black},} and $K=25$.
\section{Experimental Results}
\label{sec:experimental_results}
{\color{black}
\subsection{Datasets}

In this work, we evaluate the proposed framework on two different cross-modality re-identification datasets, SYSU-MM01 \cite{wu2017rgb} and RegDB \cite{nguyen2017person}. Additionally, we perform experiments on Market-1501 \cite{zheng2015scalable} to expose the performance of the framework for re-identification in the visible domain.

}
\subsubsection{SYSU-MM01}
\begin{figure}
    \centering
    \includegraphics[width=7 cm]{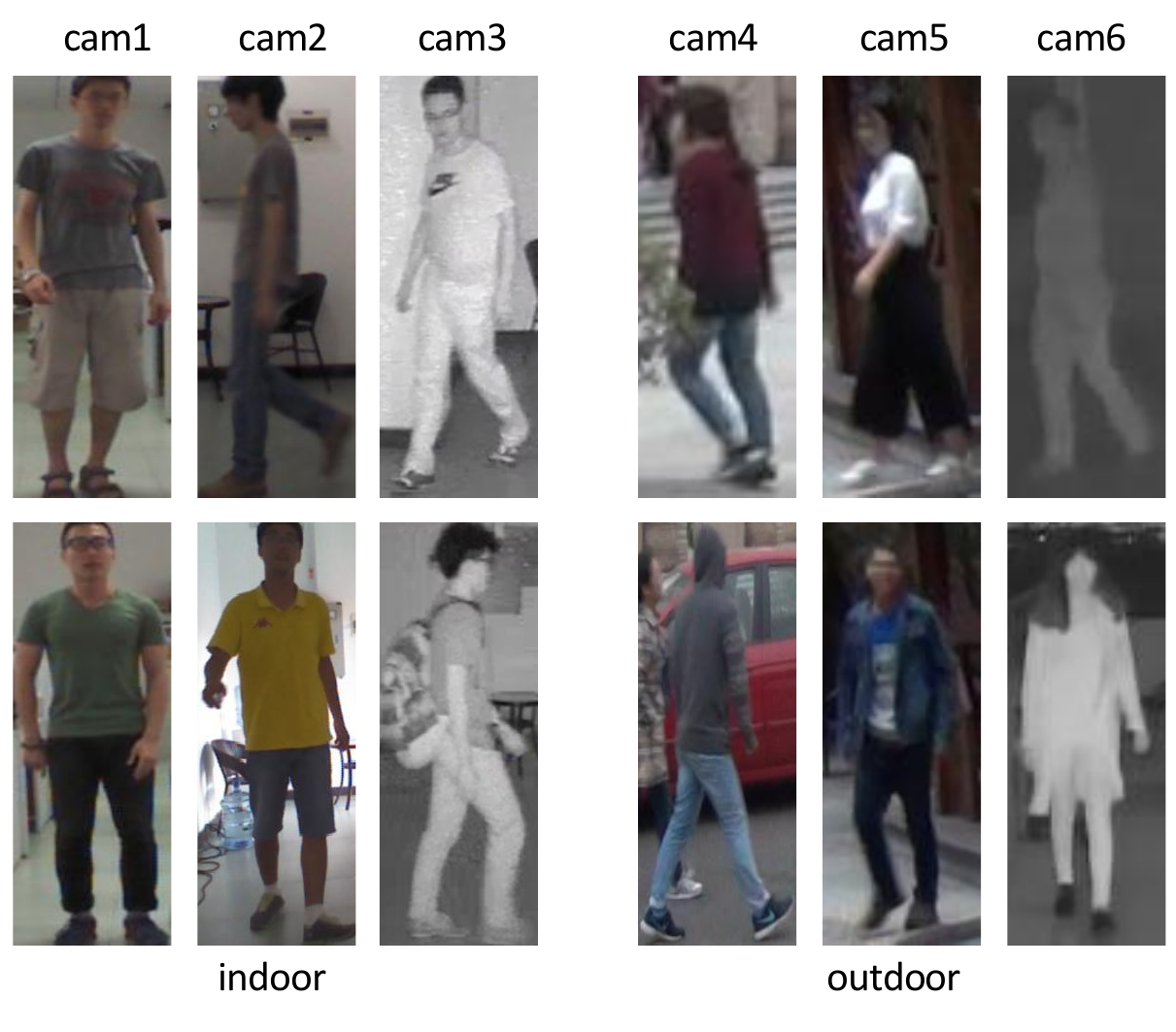}
    \caption{Sample images from SYSU-MM01 dataset.}
    \label{fig:dataset}
\end{figure}

The SYSU-MM01 \cite{wu2017rgb} is a dataset collected for VI-ReId problem. 
The images in the dataset were obtained from $491$ different persons by recording them using {\color{black}four} RGB and  {\color{black}two} infrared cameras. Within the dataset, the persons are divided into {\color{black}three} fixed splits to create training, validation{\color{black},} and test sets. 
In the training set, there are $20284$ RGB and $9929$ infrared images of $296$ persons. The validation set contains $1974$ RGB and $1980$ infrared images of $99$ persons. 
The testing set consists of the images of $96$ persons where $3803$ infrared images are used as query and $301$ randomly selected RGB images are used as gallery.
For the evaluation, there are two search modes, all-search and indoor-search. In the all-search mode, the gallery set consists of the images taken from RGB cameras (cam1, cam2, cam4, cam5), and the probe set consists of the images taken from infrared cameras (cam3 and cam6). In the indoor-search mode, the gallery set contains the RGB images only taken from the cameras (cam1 and cam2) located in the indoor. 
In the standard evaluation protocol of the dataset, the performance is reported using cumulative matching characteristic (CMC) curves and mean average precision (mAP). CMC and mAP are calculated with 10 random split{\color{black}s} of 
the gallery and probe sets, and the average performance is reported. We follow the single-shot setting for all the experiments. In Figure \ref{fig:dataset}, some sample images from SYSU-MM01 are shown.

{\color{black}
\subsubsection{RegDB}

RegDB \cite{nguyen2017person} consists of images captured using RGB and thermal dual-camera system. There are 412 persons in the dataset and each person has 20 images, 10 RGB and 10 thermal. We perform the experiments on this dataset by following the evaluation protocol proposed by Ye \textit{et al.} \cite{ye2018hierarchical}. In this protocol, the training and the test sets are created by randomly splitting the dataset into two parts. The gallery set is then created with thermal images, and the query set is created with RGB images. The results are reported with mAP and CMC, and the splitting process is repeated for 10 trials to ensure that the results are statistically stable. Sample images from RegDB are given in Figure \ref{fig:dataset_regdb}.

\begin{figure}
    \centering
    \includegraphics[width=7 cm, height=5 cm]{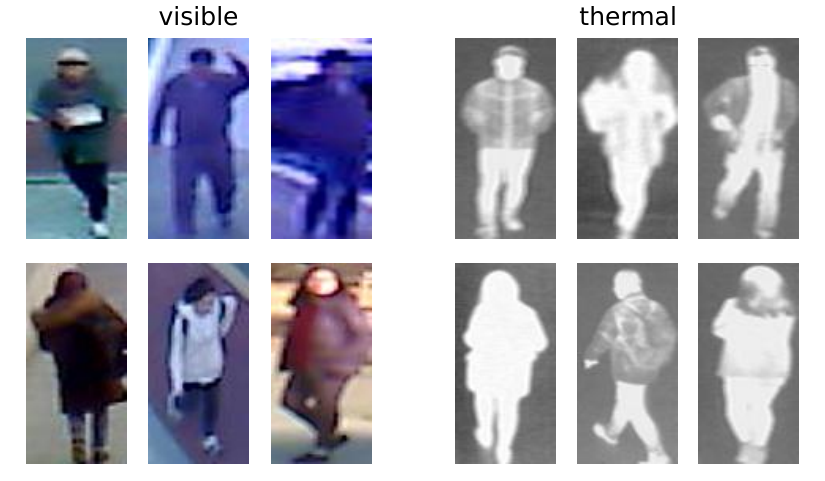}
    \caption{Sample images from RegDB dataset.}
    \label{fig:dataset_regdb}
\end{figure}

\subsubsection{Market-1501}

Market-1501 dataset \cite{zheng2015scalable} has 32668 images captured from 1501 persons using one low-resolution and five high-resolution cameras. In the standard evaluation protocol of Market-1501, 12936 images of 751 persons are reserved for the training operations. The images of the remaining 750 persons constitute the query and gallery sets with 3368 and 19732 images, respectively. CMC and mAP metrics are used to report the performance on this dataset, as in SYSU-MM01 and RegDB.

}

\subsection{Training the Networks}
\label{sec:training_nets}
While training the ResNet networks in the proposed framework, we set the resolution of all RGB, infrared{\color{black}, and thermal} input images to $492\times164$. In the experiments where the LZM transformation is used, we normalize the images to have zero mean.
{\color{black}
For SYSU-MM01 and Market-1501, we perform the training operations using the own training set of each dataset and set the maximum number of iterations to 100K. However, for RegDB, we follow a two-step training strategy similar to \cite{kalayeh2018human} to avoid overfitting. Since the training set of RegDB has a limited number of images, models easily over-fit the training set. In the first step, the networks are trained for 100K iterations using a new set created by combining the training sets of SYSU-MM01 and RegDB. In the second step, the networks are fine-tuned for 10K iterations using only the training set of RegDB.
}
{\color{black}For all the training operations (including the fine-tuning on RegDB) performed in this work, we use Nesterov Accelerated Gradient \cite{bengio2013advances} and set the mini-batch size to $8$, momentum to $0.9$ and weight decay to $0.0005$. We choose $0.01$ as the initial value of the learning rate and decay it 10 times using the exponential shift with the rate of $0.9$, such that}
\begin{equation}
\label{eqn:lr_up}
\begin{aligned}
lr\_new=lr\_init\times(1-\frac{iter}{max\_iter})^{0.9}.
\end{aligned}
\end{equation}
{\color{black}Here,} $lr\_init$ and $lr\_new$ are the initial and the updated learning rates, respectively.  $iter$ is the current number of iterations and $max\_iter$ is the maximum number of iterations.

\subsection{Re-Identification Performance {\color{black}on SYSU-MM01}}
In this section, we call the ResNet models in the streams of the proposed framework as $RX_{gray\_ir}$, $RX_{rgb\_ir}$, $RX_{LZM(gray\_ir)}${\color{black},} and $RX_{LZM(rgb\_ir)}$, respectively, to avoid confusion while reporting the experimental results. The $X$ in the labels indicates which ResNet model is used and takes one of the values 50, 101{\color{black},} and 152. In this section, we show the experimental results under all-search mode. 

As mentioned in Section \ref{sec:zernike}, the number of filters used in the LZM transformation depends on the moment degree $n$ (Eq. \ref{eqn:kn}). Therefore, we have first carried out experiments with model $R50_{LZM(gray\_ir)}$ in order to determine the optimal values for both the moment degree $n$ and the filter size $k$. The results obtained with different parameter sets are given in Table \ref{table:lzm_params}. According to this table, the results are close to each other, but slightly better results are obtained when $3/5$ and $3/7$ are used for {\color{black}the} $n/k$ pair. For this reason, we prefer to use $n=3$ and $k=5$ to generate the LZM pattern maps {\color{black}for} the rest of the study. 
\begin{table}[]
\scriptsize
\centering
\caption{Results {\color{black}on SYSU-MM01 dataset} obtained with different values of $n$ and $k$ parameters of the LZM transformation.}
\begin{tabular}{lllllll}
\hline 
\multicolumn{1}{c}{\textbf{Method}} & \textbf{n}& \textbf{k} & \textbf{Rank1} & \textbf{Rank10} & \textbf{Rank20} & \textbf{mAP} \\
\hline 
 & 2 & 5 & 30.77 & 78.74 & 89.67 & 32.28\\ 
 & 3 & 3 & 29.18 & 76.14 & 88.25 & 30.79\\ 
$R50_{LZM(gray\_ir)}$ & 3 & 5 & 31.45 & \textbf{79.27} & \textbf{90.48} & 32.57 \\ 
 & 3 & 7 & \textbf{31.48} & 78.05 & 89.29 & \textbf{32.64}\\ 
 & 4 & 5 & 30.35 & 78.79 & 89.93 & 31.92\\ 
 & 5 & 5 & 29.23 & 75.97 & 88.39 & 31.06\\ \hline
\end{tabular}
\label{table:lzm_params}
\end{table}


Table \ref{table:r50_random} shows the results obtained by training a ResNet-50 in each stream of the proposed framework. According to the results, better performance is achieved with $R50_{gray\_ir}$ and $R50_{LZM(gray\_ir)}$, where grayscale \& infrared input images are used, compared to those of $R50_{rgb\_ir}$ and $R50_{LZM(rgb\_ir)}$ using RGB \& infrared input images. Reducing the visual differences between infrared and RGB images by converting RGB images to grayscale makes a positive contribution to the performance. When the features obtained 
from the first two streams and the last two streams are combined, a significant performance increase is observed{\color{black},} as can be seen in {\color{black}the third} and  {\color{black}sixth} rows of Table \ref{table:r50_random}. With the concatenation of the features extracted by $R50_{gray\_ir}$ and $R50_{rgb\_ir}$, 4.5\% improvement is achieved for Rank-1 as compared to $R50_{gray\_ir}$. Likewise, the concatenation of the features extracted by $R50_{LZM(gray\_ir)}$ and $R50_{LZM(rgb\_ir)}$ improves Rank1 by 3.3\% as compared to $R50_{LZM(gray\_ir)}$. This shows that we generate complementary features with the models that are trained with grayscale \& infrared and RGB \& infrared input images. In the last row of Table \ref{table:r50_random}, the results achieved by combining the features obtained in all streams of the proposed framework are given. When they are compared to those of $R50_{gray\_ir}+R50_{rgb\_ir}$ and $R50_{LZM(gray\_ir)}+R50_{LZM(rgb\_ir)}$, it is seen that the improvement obtained for Rank-1 is 5.3\% and 3.7\%, respectively. Therefore, taking this performance improvement into account, we can conclude that the models trained with the LZM pattern maps generate features that are different from and  complementary to the ones produced by $R50_{gray\_ir}$ and $R50_{rgb\_ir}$.

\begin{table}[]
\scriptsize
\centering
\caption{Results {\color{black}on SYSU-MM01 dataset} obtained when ResNet-50 initialized with scaled Gaussian distribution \cite{he2015delving} is used in all the streams of the framework.}
\begin{tabular}{lccccc}
\hline
\multicolumn{1}{c}{\textbf{Method}} & \textbf{Rank1} & \textbf{Rank10} & \textbf{Rank20} & \textbf{mAP} \\ \hline 
$R50_{gray\_ir}$  & 28.59 & 75.95 & 87.57 & 30.42 \\ 
$R50_{rgb\_ir}$ & 26.20 & 70.53 & 83.24 & 28.48 \\ 
$R50_{gray\_ir}+R50_{rgb\_ir}$ & 33.06 & 79.67 & 89.87 & 35.11 \\ \hline 
$R50_{LZM(gray\_ir)}$ & 31.45 & 79.27 & 90.48 & 32.57 \\ 
$R50_{LZM(rgb\_ir)}$ & 28.32 & 73.09 & 84.98 & 30.44 \\ 
$R50_{LZM(gray\_ir)}+R50_{LZM(rgb\_ir)}$ & 34.73 & 81.13 & 91.26 & 36.30 \\ \hline 
\textbf{$All$} & \textbf{38.39} & \textbf{83.75} & \textbf{92.47} & \textbf{39.67} \\ \hline
\end{tabular}
\label{table:r50_random}
\end{table}


While performing the experiments whose results are given in Table \ref{table:lzm_params} and \ref{table:r50_random}, we have initialized the weights of ResNet-50 with scaled Gaussian distribution proposed in \cite{he2015delving}. However, many ReId studies \cite{kalayeh2018human, chen2017person, sarfraz2017pose} in the literature use pre-trained models instead of the models with randomly initialized weights. The vast majority of these pre-trained models are trained using ImageNet \cite{russakovsky2015imagenet}{\color{black},} which is a large-scale image classification dataset. In this way, the information of the model capable  {\color{black}of} image classification is utilized for ReId. By following the same approach,  we have conducted experiments using ResNet-50 {\color{black}models, which are} pre-trained on ImageNet{\color{black},} for each stream of our framework. The results are given in Table \ref{table:r50_imagenet}. In the  {\color{black}third} and  {\color{black}fourth} streams of the framework, the number of channels of the input images  {\color{black}is} greater than  {\color{black}three} because the LZM pattern maps are used. Therefore, in the experiments, we remove the first convolutional layer (conv1) of the pre-trained ResNet-50{\color{black}. Then, we}  insert a randomly initialized convolutional layer whose number of input channels matches the number of LZM pattern maps. For the first stream of the framework, we create  {\color{black}three-}channel input images by repeating the grayscale/infrared images for R, G{\color{black},} and B channels. If the results given in Table{\color{black}s} \ref{table:r50_random} and \ref{table:r50_imagenet} are compared, it is seen that we achieve a significant performance improvement for each stream by using pre-trained ResNet-50 models. With the concatenation of the features from  {\color{black}four} streams, we improve Rank-1 and mAP by 6.6\% and 6.3\% as compared to the results in the last row of Table \ref{table:r50_random} and reach to 45\% and 45.9\%, respectively.

\begin{table}[]
\scriptsize
\centering
\caption{Results {\color{black}on SYSU-MM01 dataset} obtained when {\color{black}pre-trained ResNet-50} is used in all the streams of the framework.}
\begin{tabular}{lccccc}
\hline
\multicolumn{1}{c}{\textbf{Method}} & \textbf{Rank1} & \textbf{Rank10} & \textbf{Rank20} & \textbf{mAP} \\ \hline
$R50_{gray\_ir}$  & 38.94 & 85.47 & 93.99 & 39.63 \\ 
$R50_{rgb\_ir}$ & 33.87 & 76.99 & 88.66 & 35.21 \\ 
$R50_{gray\_ir}+R50_{rgb\_ir}$ & 42.58 & 86.61 & 94.54 & 43.38 \\ \hline 
$R50_{LZM(gray\_ir)}$ & 37.30 & 84.56 & 93.60 & 37.93 \\ 
$R50_{LZM(rgb\_ir)}$ & 35.09 & 80.11 & 91.18 & 36.67 \\ 
$R50_{LZM(gray\_ir)}+R50_{LZM(rgb\_ir)}$ & 41.50 & 86.44 & 94.73 & 42.33 \\ \hline 
\textbf{$All$} & \textbf{45.00} & \textbf{89.06} & \textbf{95.77} & \textbf{45.94} \\ \hline
\end{tabular}
\label{table:r50_imagenet}
\end{table}


According to recent research on computer vision tasks such as image classification, it is observed that deeper networks are more accurate \cite{Wang_2018_ECCV}. In order to show the performance of our framework with deeper networks, we have conducted experiments using ResNet-101 and ResNet-152{\color{black},} which have more layers than ResNet-50.
In the experiments, we have used the models pre-trained on ImageNet, and we show the results in Tables \ref{table:r101_imagenet} and \ref{table:r152_imagenet}. 
As can be seen in {\color{black}the third} and {\color{black}sixth} rows of  Tables \ref{table:r101_imagenet} and \ref{table:r152_imagenet}, significant performance improvement is obtained with the concatenation of the features generated in the first two streams and in the last two streams,
as in Tables \ref{table:r50_random} and \ref{table:r50_imagenet}. The highest results are achieved when the features from all the streams are combined. Using ResNet-101 and ResNet-152, we improve the best Rank1 result given in Table \ref{table:r50_imagenet} (last row) by 1.8\% and 2.4\% and reach to 46.80\% and 47.35\%, respectively. 

\begin{table}[]
\scriptsize
\centering
\caption{Results {\color{black}on SYSU-MM01 dataset} obtained when {\color{black}pre-trained ResNet-101} is used in all the streams of the framework.}
\begin{tabular}{lccccc}
\hline
\multicolumn{1}{c}{\textbf{Method}} & \textbf{Rank1} & \textbf{Rank10} & \textbf{Rank20} & \textbf{mAP} \\ \hline
$R101_{gray\_ir}$  & 40.41 & 85.02 & 93.80 & 41.21 \\
$R101_{rgb\_ir}$ & 35.71 & 79.34 & 89.69 & 38.01 \\
$R101_{gray\_ir}+R101_{rgb\_ir}$ & 43.70 & 86.89 & 94.71 & 45.35 \\ \hline
$R101_{LZM(gray\_ir)}$ & 39.98 & 86.39 & 94.88 & 40.94 \\ 
$R101_{LZM(rgb\_ir)}$ & 37.41 & 81.92 & 92.17 & 39.25 \\ 
$R101_{LZM(gray\_ir)}+R101_{LZM(rgb\_ir)}$ & 43.58 & 88.35 & 96.02 & 44.94 \\ \hline
$All$ & \textbf{46.80} & \textbf{89.99} & \textbf{96.62} & \textbf{48.21} \\ \hline
\end{tabular}
\label{table:r101_imagenet}
\end{table}


\begin{table}[]
\scriptsize
\centering
\caption{Results {\color{black}on SYSU-MM01 dataset} obtained when  {\color{black}pre-trained ResNet-152} is used in all the streams of the framework.}
\begin{tabular}{lccccc}
\hline
\multicolumn{1}{c}{\textbf{Method}} & \textbf{Rank1} & \textbf{Rank10} & \textbf{Rank20} & \textbf{mAP} \\ \hline
$R152_{gray\_ir}$  & 38.88 & 85.01 & 93.80 & 40.23 \\
$R152_{rgb\_ir}$  & 36.27 & 78.63 & 88.74 & 38.33 \\ 
$R152_{gray\_ir}+R152_{rgb\_ir}$  & 43.54 & 86.49 & 94.40 & 45.30 \\ \hline
$R152_{LZM(gray\_ir)}$  & 40.68 & 84.89 & 93.32 & 41.02 \\
$R152_{LZM(rgb\_ir)}$  & 39.65 & 81.69 & 91.36 & 40.49 \\
$R152_{LZM(gray\_ir)}+R152_{LZM(rgb\_ir)}$  & 44.69 & 87.00 & 94.71 & 45.21 \\ \hline
$All$  & \textbf{47.35} & \textbf{89.10} & \textbf{95.67} & \textbf{48.32} \\ \hline
\end{tabular}
\label{table:r152_imagenet}
\end{table}


With the results given so far, we have demonstrated that the performance is significantly boosted by using the features learned from the LZM pattern maps.
This shows that the models trained with the LZM pattern maps generate features  {\color{black}that} are different from and complementary to the ones generated by the other models.
To further show the contribution of the features extracted from the LZM pattern maps, we have performed additional experiments and give the results in Table \ref{table:diff_comb}. 
In this table, $RX_{\{g+r\}}$ and $RX_{\{LZM(g+r)\}}$  represent the concatenation of the features of the first two  and the last two streams, such that $RX_{\{g+r\}}=[RX_{gray\_ir}, RX_{rgb\_ir}]$ and $RX_{\{LZM(g+r)\}}=[RX_{LZM(gray\_ir)},$ $RX_{LZM(rgb\_ir)}]$.
The first row of Table \ref{table:diff_comb} shows the results obtained using ResNet-50 for each stream of the proposed framework. We have computed the results given in the second row of the table by using $R101_{gray\_ir}$ and $R101_{rgb\_ir}$ for the third and fourth streams of the framework instead of ResNet-50 models that utilize the LZM pattern maps. 
In this way, there is a 0.8\% improvement for Rank-1. However, the improvement becomes 1.7\% when $R101_{gray\_ir}$ and $R101_{rgb\_ir}$ are replaced with $R101_{LZM(gray\_ir)}$ and $R101_{LZM(rgb\_ir)}${\color{black},} as shown in the third row.
Similarly, better results are achieved when $R152_{LZM(gray\_ir)}$ and $R152_{LZM(rgb\_ir)}$ are used for the third and fourth streams compared to using $R152_{gray\_ir}$ and $R152_{rgb\_ir}${\color{black},} as shown in {\color{black}the} fourth and fifth rows. The results of the experiments{\color{black},} where the ResNet-50 models are used for the first and second streams of the framework{\color{black},} are given in the first five rows of Table \ref{table:diff_comb}. 
The next  {\color{black}five} rows and the last  {\color{black}five} rows show the results obtained using ResNet-101 and ResNet-152 for the first two streams, respectively. 
Like the results given in the first  {\color{black}five} rows, these results also demonstrate that the LZM transformation plays an important role.
Using the models trained with the LZM pattern maps, better performance improvements are achieved compared to the other models with the same depth and trained with grayscale \& infrared and RGB \& infrared images. This verifies that, by exposing the texture information from the images, the LZM transformation enables the ResNet architectures to learn different {\color{black}as well as} complementary features.


\begin{table}[]
\scriptsize
\centering
\caption{Results {\color{black}on SYSU-MM01 dataset} obtained using ResNet-50, ResNet-101 and ResNet-152 models in different combinations. $RX_{\{g+r\}}$ and $RX_{\{LZM(g+r)\}}$  represent the concatenation of the features from the first and second, and third and fourth streams, respectively.}
\begin{tabular}{lcccc}
\hline
\multicolumn{1}{c}{\textbf{Method}} & \textbf{Rank1} & \textbf{Rank10} & \textbf{Rank20} & \textbf{mAP} \\ \hline
$R50_{\{g+r\}}+R50_{\{LZM(g+r)\}}$  & 45.00 & 89.06 & 95.77 & 45.94 \\
$R50_{\{g+r\}}+R101_{\{g+r\}}$ & 45.80 & 88.51 & 95.53 & 46.94 \\
$R50_{\{g+r\}}+R101_{\{LZM(g+r)\}}$ & 46.69 & \textbf{90.32} & 96.59 & 47.79 \\
$R50_{\{g+r\}}+R152_{\{g+r\}}$ & 46.02 & 88.25 & 95.44 & 47.17 \\
$R50_{\{g+r\}}+R152_{\{LZM(g+r)\}}$ & \textbf{47.51} & 89.59 & 96.01 & 48.15 \\ \hline
$R101_{\{g+r\}}+R50_{\{g+r\}}$ & 45.80 & 88.51 & 95.53 & 46.94 \\
$R101_{\{g+r\}}+R50_{\{LZM(g+r)\}}$ & 46.23 & 89.22 & 96.11 & 47.42 \\
$R101_{\{g+r\}}+R101_{\{LZM(g+r)\}}$ & 46.80 & 89.99 & \textbf{96.62} & 48.21 \\
$R101_{\{g+r\}}+R152_{\{g+r\}}$ & 45.84 & 88.13 & 95.34 & 47.50 \\ 
$R101_{\{g+r\}}+R152_{\{LZM(g+r)\}}$ & 47.24 & 89.12 & 96.09 & 48.27 \\ \hline 
$R152_{\{g+r\}}+R50_{\{g+r\}}$ & 46.02 & 88.25 & 95.44 & 47.17 \\ 
$R152_{\{g+r\}}+R50_{\{LZM(g+r)\}}$ & 46.57 & 89.16 & 95.97 & 47.75 \\ 
$R152_{\{g+r\}}+R101_{\{g+r\}}$ & 45.84 & 88.13 & 95.34 & 47.50 \\ 
$R152_{\{g+r\}}+R101_{\{LZM(g+r)\}}$ & 47.01 & 89.63 & 96.33 & \textbf{48.37} \\ 
\textbf{$R152_{\{g+r\}}+R152_{\{LZM(g+r)\}}$} & 47.35 & 89.10 & 95.67 & 48.32 \\ \hline
\end{tabular}
\label{table:diff_comb}
\end{table}


\subsubsection{Comparison With the State-Of-the-Art}

Tables \ref{table:soa_all} and \ref{table:soa_in} show our results in comparison with the state-of-the-art. 
In \cite{wu2017rgb}, zero-padding is utilized to enable a one-stream network to learn domain-specific structures automatically. TONE+HCML \cite{ye2018hierarchical} and TONE+XQDA \cite{ye2018hierarchical} use a two-stage framework that includes feature learning and metric learning. BCTR \cite{ye2018visible} and BDTR \cite{ye2018visible} have a two-stream framework that employs separate networks for RGB and infrared images to extract domain-specific features. 
Kang \textit{et al.} \cite{kang2019person} train a one-stream network using single input images generated by placing visible and infrared images in different channels or by concatenating them.
In \cite{dai2018cross}, a generative adversarial network is used in order to extract common features for the images in different domains. 
D$^2$RL \cite{wanglearning} train a one-stream network with multi-spectral images generated by projecting RGB and infrared images to a unified space.
Different  {\color{black}from} these works, in this study, we train multiple ResNet architectures by using the different representation{\color{black}s} of the input images and generate complementary features from each one of them for RGB and infrared images. As can be seen from Tables \ref{table:soa_all} and \ref{table:soa_in}, we outperform the current state-of-the-art with a large margin. 
When we use ResNet-152 architectures in {\color{black}all} the  streams, our framework improves VI-ReId performance on SYSU-MM01 under all-search mode by $14.09\%/12.10\%$ and under indoor-search mode by $14.48\%/16.01\%$ in Rank-1$/$mAP. 
By concatenating all the feature vectors generated with ResNet-50, ResNet-101{\color{black},} and ResNet-152, we obtain at least 1.5\% additional improvements for Rank-1 and mAP under both search modes.
When we perform the re-ranking \cite{sarfraz2017pose}, the margin of the improvement further increases. We achieve $63.05\%/67.13\%$ and $69.06\%/76.95\%$ under all-search and indoor-search modes, respectively.
\begin{table}[h]
\scriptsize
\centering
\caption{State-of-the-art comparison {\color{black}on SYSU-MM01 dataset} under all-search mode.}
\begin{tabular}{lcccc}
\hline
\multicolumn{1}{c}{\textbf{Method}} & \textbf{Rank1} & \textbf{Rank10} & \textbf{Rank20} & \textbf{mAP}
\\ \hline 
Lin \textit{et al.}* \cite{lin2017cross} & 5.29 & 33.71 & 52.95 & 8.00
\\ 
One-stream \cite{wu2017rgb} & 12.04 & 49.68 & 66.74 & 13.67
\\ 
Two-stream \cite{wu2017rgb} & 11.65 & 47.99 & 65.50 & 12.85
\\ 
Zero-padding \cite{wu2017rgb} & 14.80 & 54.12 & 71.33 & 15.95
\\ 
TONE+XQDA* \cite{ye2018hierarchical} & 14.01 & 52.78 & 69.06 & 15.97
\\ 
TONE+HCML* \cite{ye2018hierarchical} & 14.32 & 53.16 & 69.17 & 16.16
\\ 
BCTR \cite{ye2018visible} & 16.12 & 54.90 & 71.47 & 19.15
\\ 
BDTR \cite{ye2018visible} & 17.01 & 55.43 & 71.96 & 19.66 
\\ 
Kang \textit{et al.} \cite{kang2019person} & 23.18 & 51.21 & 61.73 & 22.49
\\ 
cmGAN \cite{dai2018cross} & 26.97 & 67.51 & 80.56 & 27.80
\\
{\color{black}eBDTR(ResNet50) \cite{ye2019bi}} & 27.82 & 67.34 & 81.34 & 28.42
\\
D$^2$RL \cite{wanglearning} & 28.90 & 70.60 & 82.40 & 29.20
\\
{\color{black}Ye \textit{et al.} \cite{ye2019improving}} & 31.41 & 73.75 & 86.29 & 33.18 
\\
{\color{black}MAC \cite{ye2019modality}} & 33.26 & 79.04 & 90.09 & 36.22 
\\
\hline 
$R50_{\{g+r\}}+R50_{\{LZM(g+r)\}}$  & 45.00 & 89.06 & 95.77 & 45.94
\\ 
$R101_{\{g+r\}}+R101_{\{LZM(g+r)\}}$ & 46.80 & 89.99 & 96.62 & 48.21 
\\ 
$R152_{\{g+r\}}+R152_{\{LZM(g+r)\}}$ & 47.35 & 89.10 & 95.67 & 48.32
\\ 
All & 48.87 & 90.73 & \textbf{96.72} & 49.85
\\ 
All + re-ranking & \textbf{63.05} & \textbf{93.62} & 96.30 & \textbf{67.13}
\\ \hline
\multicolumn{5}{l}{* indicates the results copied from \cite{ye2018visible}.}
\end{tabular}
\label{table:soa_all}
\end{table}
\begin{table}[]
\scriptsize
\centering
\caption{State-of-the-art comparison {\color{black}on SYSU-MM01 dataset} under indoor-search mode.}
\begin{tabular}{lcccc}
\hline
\multicolumn{1}{c}{\textbf{Method}} & \textbf{Rank1} & \textbf{Rank10} & \textbf{Rank20} & \textbf{mAP}
\\ \hline 
Lin \textit{et al.}* \cite{lin2017cross} & 9.46 & 48.98 & 72.06 & 15.57
\\ 
One-stream \cite{wu2017rgb} & 16.94 & 63.55 & 82.10 & 22.95
\\ 
Two-stream \cite{wu2017rgb} & 15.60 & 61.18 & 81.02 & 21.49
\\ 
Zero-padding \cite{wu2017rgb} & 20.58 & 68.38 & 85.79 & 26.92 \\ 
cmGAN \cite{dai2018cross} & 31.63 & 77.23 & 89.18 & 42.19
\\ 
{\color{black}eBDTR(ResNet50) \cite{ye2019bi}} & 32.46 & 77.42 & 89.62 & 42.46
\\
{\color{black}MAC \cite{ye2019modality}} & 33.37 & 82.49 & 93.69 & 44.95 
\\
{\color{black}Ye \textit{et al.} \cite{ye2019improving}} & 37.62 & 83.27 & 93.56 & 46.32
\\
\hline 
$R50_{\{g+r\}}+R50_{\{LZM(g+r)\}}$  & 49.66 & 92.47 & 97.15 & 59.81
\\ 
$R101_{\{g+r\}}+R101_{\{LZM(g+r)\}}$ & 53.01 & 94.05 & \textbf{98.44} & 62.86
\\ 
$R152_{\{g+r\}}+R152_{\{LZM(g+r)\}}$ & 52.10 & 93.69 & 98.06 & 62.33
\\ 
All & 54.28 & 94.22 & 98.22 & 63.92
\\ 
All + re-ranking & \textbf{69.06} & \textbf{96.30} & 97.16 & \textbf{76.95}
\\ \hline
\multicolumn{5}{l}{* indicates the results copied from \cite{wu2017rgb}.}
\end{tabular}
\label{table:soa_in}
\end{table}

{\color{black}
\subsection{Re-Identification Performance on RegDB}

In the experiments on the RegDB dataset, we employ only the ResNet-50 models due to the relatively low number of training images. As noted in Section \ref{sec:training_nets}, we train these models in two steps. In the first step, the training is carried out using a set created by combining SYSU-MM01 and RegDB training sets. Then, in the second step, the models are fine-tuned using only the images from the RegDB dataset. 
With such a training strategy, we aim to avoid overfitting.

According to the results given in Table \ref{table:regdb_v2t}, the models trained with LZM pattern maps exhibit lower performance than the others. As shown in Figures \ref{fig:filters} and \ref{fig:lzm_outs}, LZM filters used perform high-pass filtering and reduce the low-frequency components of the images. This reduction results in LZM pattern maps to have less information compared to the original images. 
For this reason, in order to learn features with the same level of discriminating information (compared to using RGB or grayscale images), more training images will be needed in the training process performed with LZM pattern maps.
In the first phase of the two-step training strategy, we expand the training set using the images from the SYSU-MM01 database. However, the images in the SYSU-MM01 and RegDB databases were recorded 
at different domains and so have different features from each other.
Therefore, the models trained with the LZM pattern maps are not fed as many different images as they need in order to learn the specific features of the images in RegDB. 
As a result, models trained with the LZM pattern maps have lower performance than others.
On the other hand, when the features obtained in all the streams of the proposed framework are combined, it is observed that Rank-1 and mAP are improved by 1.91\% and 1.84\%, respectively. This demonstrates that different and complementary features can be learned for also the RegDB dataset by using LZM pattern maps. 
Similar to \cite{ye2018hierarchical}, we have conducted additional experiments to evaluate the performance of our framework by following a different gallery/query setting where the RGB images are used as the gallery and the thermal images used as the query set.
The results are given in Table \ref{table:regdb_t2v}. It is observed that the features of the models $R50_{LZM(rgb\_ir)}$ and $R50_{LZM(gray\_ir)}$ contribute to the performance by improving the Rank-1 and mAP by 2.32\% and 2.22\%, respectively.


\begin{table}[]
\scriptsize
\centering
\caption{{\color{black}Results on RegDB dataset under visible to thermal setting.}}
\begin{tabular}{>{\color{black}}l>{\color{black}}c>{\color{black}}c>{\color{black}}c>{\color{black}}c>{\color{black}}c}
\hline
\multicolumn{1}{>{\color{black}}c}{\textbf{Method}} & \textbf{Rank1} & \textbf{Rank10} & \textbf{Rank20} & \textbf{mAP} \\ \hline
$R50_{gray\_ir}$  & 43.94 & 67.28 & 77.50 & 45.28 \\ 
$R50_{rgb\_ir}$ & 49.48 & 69.60 & 79.30 & 50.42 \\ 
$R50_{gray\_ir}+R50_{rgb\_ir}$ & 55.12 & 75.37 & 83.85 & 56.22 \\  \hline 
$R50_{LZM(gray\_ir)}$ & 38.78 & 61.15 & 71.54 & 40.14 \\ 
$R50_{LZM(rgb\_ir)}$ & 36.85 & 59.25 & 70.39 & 38.62 \\ 
$R50_{LZM(gray\_ir)}+R50_{LZM(rgb\_ir)}$ & 44.11 & 65.94 & 75.84 & 45.70 \\  \hline 
\textbf{$All$} & \textbf{57.03} & \textbf{76.10} & \textbf{84.34} & \textbf{58.06} \\ \hline
\end{tabular}
\label{table:regdb_v2t}
\end{table}

\begin{table}[]
\scriptsize
\centering
\caption{{\color{black}Results on RegDB dataset under thermal to visible setting.}}
\begin{tabular}{>{\color{black}}l>{\color{black}}c>{\color{black}}c>{\color{black}}c>{\color{black}}c>{\color{black}}c}
\hline
\multicolumn{1}{>{\color{black}}c}{\textbf{Method}} & \textbf{Rank1} & \textbf{Rank10} & \textbf{Rank20} & \textbf{mAP} \\ \hline
$R50_{gray\_ir}$  & 45.62 & 69.13 & 78.80 & 45.51 \\ 
$R50_{rgb\_ir}$ & 48.36 & 68.57 & 78.53 & 49.19 \\ 
$R50_{gray\_ir}+R50_{rgb\_ir}$ & 54.85 & 74.05 & 82.91 & 55.34 \\  \hline 
$R50_{LZM(gray\_ir)}$ & 39.54 & 63.99 & 74.08 & 40.10 \\ 
$R50_{LZM(rgb\_ir)}$ & 37.51 & 61.61 & 71.75 & 38.13 \\ 
$R50_{LZM(gray\_ir)}+R50_{LZM(rgb\_ir)}$ & 45.18 & 67.50 & 77.57 & 45.53 \\  \hline 
\textbf{$All$} & \textbf{57.17} & \textbf{76.62} & \textbf{84.88} & \textbf{57.56} \\ \hline
\end{tabular}
\label{table:regdb_t2v}
\end{table}

\subsubsection{Comparison With the State-Of-the-Art}

In Table \ref{table:soa_regdb}, our results are compared with the state-of-the-art. It is seen that we outperform the current state-of-the-art with a large margin by improving Rank-1 and mAP by $6.18\%$ and $11.06\%$, respectively. The improvements become $9.73\%$ and $16.36\%$ when we perform the re-ranking.

\begin{table}[h]
\scriptsize
\centering
\caption{{\color{black}State-of-the-art comparison on RegDB dataset.}}
\begin{tabular}{>{\color{black}}l>{\color{black}}c>{\color{black}}c>{\color{black}}c>{\color{black}}c}
\hline
\multicolumn{1}{>{\color{black}}c}{\textbf{Method}} & \textbf{Rank1} & \textbf{Rank10} & \textbf{Rank20} & \textbf{mAP}
\\ \hline 
Lin \textit{et al.}* \cite{lin2017cross} & 17.28 & 34.47 & 45.26 & 15.06
\\ 
One-stream* \cite{wu2017rgb} & 13.11 & 32.98 & 42.51 & 14.02
\\ 
Two-stream* \cite{wu2017rgb} & 12.43 & 30.36 & 40.96 & 13.42
\\ 
Zero-padding* \cite{wu2017rgb} & 17.75 & 34.21 & 44.35 & 18.90
\\ 
TONE+XQDA* \cite{ye2018hierarchical} & 21.94 & 45.05 & 55.73 & 21.80
\\ 
TONE+HCML \cite{ye2018hierarchical} & 24.44 & 47.53 & 56.78 & 20.80
\\ 
BCTR \cite{ye2018visible} & 32.67 & 57.64 & 66.58 & 30.99
\\ 
BDTR \cite{ye2018visible} & 33.47 & 58.42 & 67.52 & 31.83
\\ 
eBDTR (AlexNet) \cite{ye2019bi} & 34.62 & 58.96 & 68.72 & 33.46
\\
Ye \textit{et al.} \cite{ye2019improving} & 35.42 & 53.75 & 64.08 & 36.42
\\
MAC \cite{ye2019modality} & 36.43 & 62.36 & 71.63 & 37.03 
\\
D$^2$RL \cite{wanglearning} & 43.40 & 66.10 & 76.30 & 44.10
\\
D-HSME \cite{hao2019hsme} & 50.85 & 73.36 & 81.66 & 47.00
\\
\hline 
$R50_{\{g+r\}}+R50_{\{LZM(g+r)\}}$  & 57.03 & \textbf{76.10} & \textbf{84.34} & 58.06
\\ 
+ re-ranking & \textbf{60.58} & 67.71 & 77.13 & \textbf{63.36}
\\ \hline
\multicolumn{5}{>{\color{black}}l}{* indicates the results copied from \cite{ye2018visible}.}
\end{tabular}
\label{table:soa_regdb}
\end{table}

\subsection{Re-Identification Performance on Market-1501}

In this section, we have performed experiments on the Market-1501 dataset to expose the performance of the LZM transformation for the ReId problem in the visible domain. In these experiments, we use the ResNet-50 model and give the results in Table \ref{table:r50_market}. The performance of ReId in the visible domain depends on the effective use of both low and high-frequency information of the images. However, as mentioned in the previous section, the LZM filters used perform high-pass filtering by reducing the low-frequency components. Therefore, lower performance is obtained with the features extracted from the LZM pattern maps. Results given in Table \ref{table:r50_market} indicate that the features of $R50_{LZM(gray\_ir)}$ and $R50_{LZM(rgb\_ir)}$ do not contribute to the performance obtained with the features of $R50_{gray\_ir}$ and $R50_{rgb\_ir}$.

\begin{table}[]
\scriptsize
\centering
\caption{{\color{black}Results on Market-1501 dataset.}}
\begin{tabular}{>{\color{black}}l>{\color{black}}c>{\color{black}}c>{\color{black}}c>{\color{black}}c>{\color{black}}c}
\hline
\multicolumn{1}{>{\color{black}}c}{\textbf{Method}} & \textbf{Rank1} & \textbf{Rank10} & \textbf{Rank20} & \textbf{mAP} \\ \hline
$R50_{gray}$  & 69.09 & 90.32 & 94.06 & 44.71 \\ 
$R50_{rgb}$ & 85.18 & 96.05 & 97.33 & 68.81 \\ 
$R50_{gray}+R50_{rgb}$ & 88.57 & \textbf{96.85} & \textbf{98.28} & \textbf{72.91} \\  \hline 
$R50_{LZM(gray)}$ & 60.96 & 87.38 & 91.30 & 35.86 \\ 
$R50_{LZM(rgb)}$ & 76.31 & 92.22 & 95.31 & 53.36 \\ 
$R50_{LZM(gray)}+R50_{LZM(rgb)}$ & 80.52 & 94.36 & 96.08 & 58.03 \\  \hline 
\textbf{$All$} & \textbf{88.78} & \textbf{96.85} & 97.83 & 71.83 \\ \hline
\end{tabular}
\label{table:r50_market}
\end{table}

\subsection{Discussion on Execution Time and Memory Utilization}

\begin{figure}
    \centering
    \includegraphics[width=.5\textwidth]{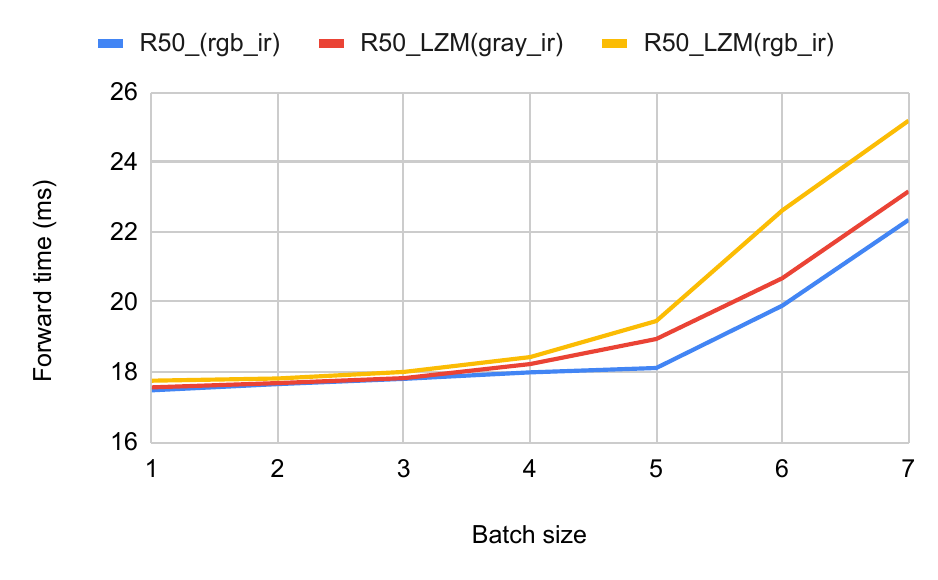}
    \caption{{\color{black}Forward pass times of $R50_{rgb\_ir}$, $R50_{LZM(gray\_ir)}$, and $R50_{LZM(rgb\_ir)}$.}}
    \label{fig:ex_comp}
\end{figure}

\begin{figure}[h]
\centering
\begin{subfigure}{.5\textwidth}
  \centering
  \includegraphics[width=1.0\linewidth]{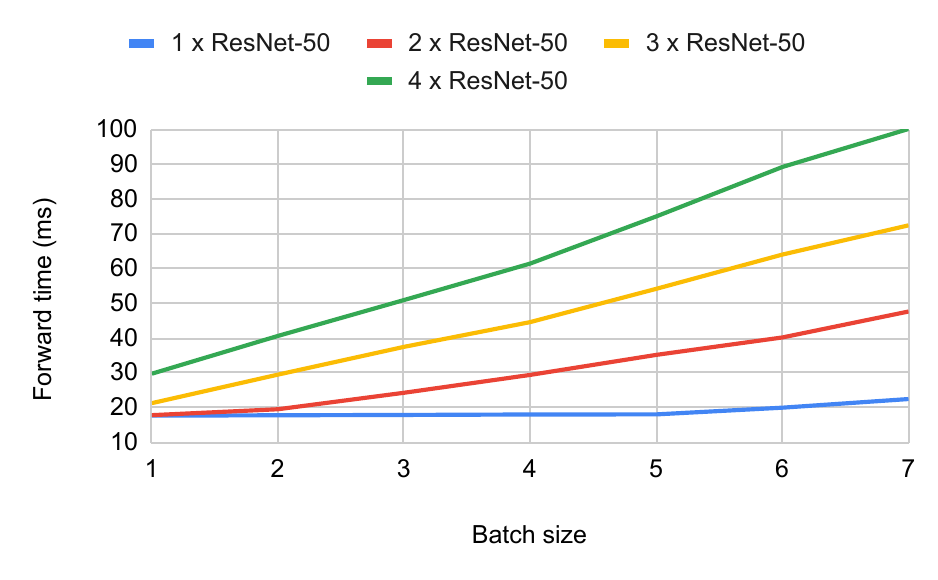}
  \caption{{\color{black}}}
  \label{fig:ex_time}
\end{subfigure}
\\
\begin{subfigure}{.5\textwidth}
  \centering
  \includegraphics[width=1.0\linewidth]{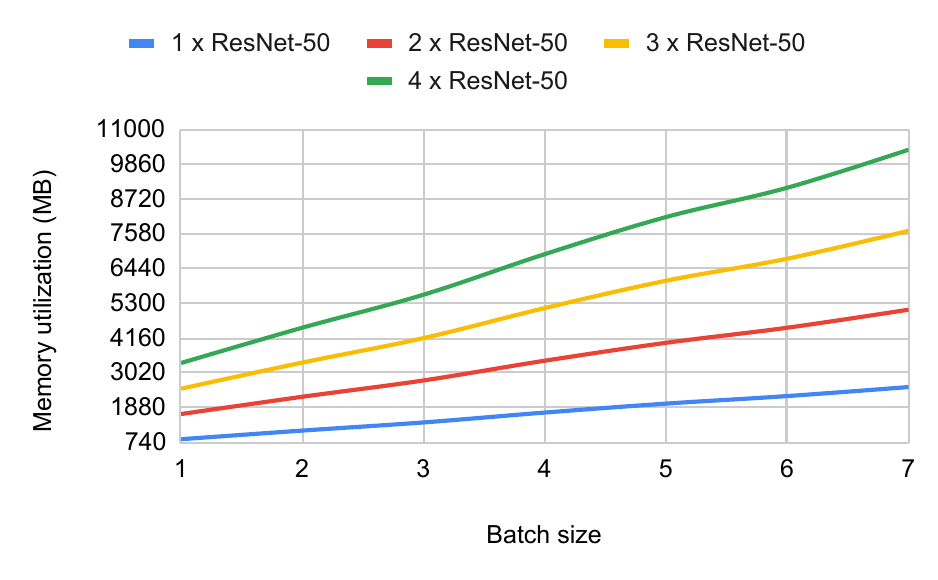}
  \caption{{\color{black}}}
  \label{fig:ex_mem}
\end{subfigure}
\caption{{\color{black}Execution time (a) and memory utilization (b) when running 1, 2, 3, and 4 ResNet-50 models together on the same GPU. Due to the lack of enough GPU memory, we can only show the results when the batch size is up to 7.}}
\label{fig:fig}
\end{figure}

We have used the Chainer framework \cite{tokui2015chainer} for the implementation of the ResNet models used in this study and carried out the experiments on a PC with an Intel Core i7-4790 CPU (3.60GHz x 8), 32 GB RAM and an Nvidia GeForce GTX 1080Ti GPU. 
We have performed all the LZM calculations on the GPU. 
In this section, we show the execution time and the GPU memory utilization of the proposed framework by using ResNet-50 models.

$RX_{gray\_ir}$ and $RX_{rgb\_ir}$ models are trained using the input images with 3 channels. For the other models, $RX_{LZM(gray\_ir)}$ and $RX_{LZM(rgb\_ir)}$, the input images have 8 and 24 channels, respectively. Therefore, in this section, we initially compare the forward pass times of $R50_{rgb\_ir}$, $R50_{LZM(gray\_ir)}$, and $R50_{LZM(rgb\_ir)}$. The results are given graphically in Figure \ref{fig:ex_comp}, where the graphs of $R50_{LZM(gray\_ir)}$ and $R50_{LZM(rgb\_ir)}$ show the total time spent on LZM transformation and ResNet-50. In Figure \ref{fig:ex_comp}, it is seen that there is no significant difference in the forward pass times of the models until the batch size is four. When the batch size is five or more, the difference begins to occur. This is because the GPU utilization is less than 100\% for all the three models when the batch size is four or less. After GPU utilization reaches 100\%, delays occur in $R50_{LZM(gray\_ir)}$ and $R50_{LZM(rgb\_ir)}$ compared to the $R50_{rgb\_ir}$.

Figures \ref{fig:ex_time} and \ref{fig:ex_mem} show forward pass time and GPU memory utilization when running 1, 2, 3, and 4 ResNet-50 models together on the same GPU. The last two of these models are $R50_{LZM(gray\_ir)}$ and $R50_{LZM(rgb\_ir)}$, respectively.
According to Figure \ref{fig:ex_time}, when the batch size is one, the running of the four models increases the execution time by $1.7$ times compared to the running of a single model. This rate increases if the number of images in the batch increases. 
However, since there is no weight sharing between the models, each model can perform feature extraction independently. Therefore, the execution time can be easily pulled down using more GPUs.
When the batch size is one, a single ResNet-50 needs 827 MB of memory, and this memory requirement increases to 2541 MB when the batch size is seven. As shown in Figure \ref{fig:ex_mem}, the memory requirement is directly proportional to the number of models running together. Due to the lack of enough GPU memory, we were unable to run four models together for larger batch sizes.

}

\section{Conclusion}
\label{sec:conclusion}

In this study, we have introduced a {\color{black}four-}stream framework for VI-ReId using ResNet architectures. In each stream of the framework, we train a ResNet by using a different representation of input images in order to obtain complementary features as much as possible from each stream. While grayscale and infrared input images are used to train the ResNet in the first stream, RGB and  {\color{black}three-}channel infrared images are used in the second stream. The first stream learns the features by using only the shape and texture information. The second stream uses the color information of the RGB images and learns to extract common features for the images with visually large differences. Unlike the first two streams, the input images in the other two streams are local pattern maps generated by employing the LZM transformation. Due to the lack of color, which provides the most important cues, in infrared images, the local shape and texture information is critical for VI-ReId. In the  {\color{black}third and fourth} streams, we expose this information from the images by generating the LZM pattern maps and train the ResNets using these maps as input images. With the exhaustive experiments performed employing  {\color{black}three} different ResNet architectures with different depths, we have demonstrated that each stream extracts different and complementary features and  {\color{black}provides} a significant contribution to the performance. Our framework outperforms{\color{black}, with a large margin, the} current state-of-the-art on SYSU-MM01 {\color{black} and RegDB datasets}. We further increase the improvement margin by utilizing re-ranking.

\bibliography{mybibfile}

\end{document}